\documentclass[11pt,a4paper]{article}

\usepackage[hyperref]{eacl2021}
\usepackage{times}
\usepackage{url}
\usepackage{latexsym}
\usepackage{soul}
\usepackage{booktabs}
\usepackage{multirow}
\usepackage{siunitx}
\usepackage{graphicx}
\usepackage{amsmath,amsfonts,amsthm,bm}
\usepackage{xcolor}
\usepackage{array}
\usepackage{caption}
\usepackage{subcaption}
\usepackage{xspace}
\usepackage[table]{colortbl}

\newcommand{\system}[1]{\textsc{#1}\xspace}
\newcommand{\Atis}{\mbox{\textsc{Atis}}\xspace}
\newcommand{\Geo}{\mbox{\textsc{GeoQuery}}\xspace}
\newcommand{\Jobs}{\mbox{\textsc{Jobs}}\xspace}
\newcommand{\seqseq}{\system{Seq2Seq}}

\newcommand{\coarsefine}{\system{Coarse2Fine}}
\newcommand{\tranx}{\system{TranX}}

\newcommand{\sperturb}{S_{\text{perturb}}(\mathbf{x}, \mathbf{y})}
\newcommand{\Dperturb}{S_{\text{perturb}}(D)}
\newcommand{\Reval}{R_{\text{eval}}(D)}
\newcommand{\Battack}{B_{\text{attack}}(D)}
\newcommand*{\affmark}[1][*]{\textsuperscript{#1}}

\newcommand\MyBox[2]{
  \fbox{\lower0.75cm
    \vbox to 1.9cm{\vfil
      \hbox to 1.9cm{\hfil\parbox{1.7cm}{#1\\#2}\hfil}
      \vfil}%
  }%
}

\aclfinalcopy 
\def\aclpaperid{1279} 
\title{On Robustness of Neural Semantic Parsers}

\author{Shuo Huang\affmark[1], Zhuang Li\affmark[1], Lizhen Qu\affmark[1]\thanks{corresponding author}\affmark[1], Lei Pan\affmark[2] \\
Faculty of Information and Technology, Monash University\affmark[1]\\
School of Info Technology, Deakin University\affmark[2]\\

{\tt shua0043@student.monash.edu}\\{\tt Zhuang.Li@monash.edu} \\ {\tt Lizhen.Qu@monash.edu} \\ {\tt l.pan@deakin.edu.au
}}

\date{}

\begin{document}

\maketitle

\begin{abstract}
Semantic parsing maps natural language (NL) utterances into logical forms (LFs), which underpins many advanced NLP problems. Semantic parsers gain performance boosts with deep neural networks, but inherit vulnerabilities against adversarial examples. In this paper, we provide the empirical study on the robustness of semantic parsers in the presence of adversarial attacks. Formally, adversaries of semantic parsing are considered to be the perturbed utterance-LF pairs, whose utterances have exactly the same meanings as the original ones. A scalable methodology is proposed to construct robustness test sets based on existing benchmark corpora. Our results answered five research questions in measuring the sate-of-the-art parsers' performance on robustness test sets, and evaluating the effect of data augmentation.

\end{abstract}

\section{Introduction}

Semantic parsing aims to map natural language (NL) utterances into logical forms (LFs), which can be executed on a knowledge base (KB) to yield denotations~\cite{Aishwarya2018SemanticParsing}. At the core of the state-of-the-art (SOTA) semantic parsers are deep learning models, which are widely known to be vulnerable to \textit{adversarial samples}~\cite{zhang2020adversarial}. This kind of examples is created by adding tiny perturbations to inputs but can severely deteriorate model performance. To the best of our knowledge, despite the popularity of semantic parsing~\cite{Aishwarya2018SemanticParsing}, there is still no published work on studying the robustness of neural semantic parsers against adversarial examples. Therefore, we conduct the \textit{first} empirical study to evaluate the effect of adversarial examples on SOTA neural semantic parsers.   

Unlike other disciplines, it is unclear what adversaries are for semantic parsers. For computer vision systems, adversaries are often generated by modifying inputs with \textit{imperceptible} perturbations. In contrast, a flip of single word or a character in an utterance can significantly change its meaning so that the changes are perceptible by humans. To address this issue,~\cite{michel2019evaluationAdverSeq2Seq} argue that adversaries for sequence-to-sequence models should maximally retain meanings after perturbing inputs. However, any meaning-changing utterances are supposed to have different meaning representations. A robust semantic parser should be invariant to meaning-preserving modifications. In light of this, given a semantic parser, we define its adversaries as the perturbed utterances satisfying two conditions: i) they have \textit{exactly} the same meanings as the original ones according to human judgements; and ii) the parser consistently produces incorrect LFs on them.    

Although new evaluation frameworks are proposed for NLP tasks~\cite{xu2020elephant,michel2019evaluationAdverSeq2Seq}, there is no framework designed for assessing robustness of semantic parsers against meaning-preserving adversaries. The current evaluation metrics focus only on standard accuracy, which measures to what degree predictions match gold standards. As pointed out by~\cite{tsipras2018robustnessOdds}, it is challenging to achieve both high standard accuracy and high robust accuracy, which measures the accuracy on adversarially perturbed examples. In order to facilitate devising novel methods for robust semantic parsers, it is desirable to develop a semantic parsing evaluation framework considering both measures.


In this work, we propose an evaluation framework for robust semantic parsing. The framework consists of an evaluation corpus and a set of customized metrics. We construct the evaluation corpus by extending three existing semantic parsing benchmark corpora. In order to generate meaning-preserving examples, we apply automatic methods to modify the utterances in those benchmark corpora by paraphrasing and injecting grammatical errors. Among the perturbed examples generated from the test sets, we build meaning-preserving test sets by filtering out the meaning-changing ones using crowdsourcing. The robustness of the semantic parsers is measured by a set of custom metrics, with and without adversarial training methods.

We conduct the \textit{first} empirical study on the robustness of semantic parsing by evaluating three SOTA neural semantic parsers using the proposed framework. The key findings from our experiments are three-folds: 
\begin{itemize}
    \item None of those SOTA semantic parsers can consistently outperform the others in terms of robustness against meaning-preserving adversarial examples; 
    \item Those neural semantic parsers are more robust to word-level perturbations than sentence-level ones; 
    \item Adversarial training through data augmentation indeed significantly improve the robustness accuracy but can only slightly influence standard accuracy.
\end{itemize}
The generated corpus and source code are available at \textcolor{darkblue}{https://github.com/shuo956/On-Robustness-of-nerual-smentic-parsing.git} 
\section{Related Work}

\paragraph{Semantic Parsing}
The SOTA neural semantic parsers formulate this task as a machine translation problem. They extend \seqseq~with attention~\cite{luong2015effective} to map NL utterances into LFs in target languages (e.g.,~lambda calculus, SQL, Python, etc.). One type of such parsers directly generates sequences of predicates as LFs~\cite{dong2016language,dong2018coarse2fine,huang2018natural}. The other type of parsers utilizes grammar rules to constrain the search space of LFs during decoding~\cite{yin2018tranx,chen2018sequence,Guo2019Towardstexttosql,rat-sql}. However, neither of the two types of parsers are evaluated against adversarial examples.

\paragraph{Adversarial Examples}
The adversarial examples are firstly defined and investigated in computer vision. Adversarial examples in that field are generated by adding imperceptible noise to input images, which lead to false predictions of machine learning models~\cite{Madry2018TowardsDL,goodfellow2014explaining}. However, it is non-trivial to add such noise to text in natural language processing (NLP) tasks due to the discrete nature of languages. Minor changes in characters or words may be perceptible to humans and may lead to change of meanings. To date, it is still difficult to reach an agreement on the definition of adversarial examples across tasks.

\newcite{Jia2017ReadingComprehens,Yonatan2017synthetic, javid2018character,miyato2016adversarial} add distracting sentences and sequences of random words into text or flip randomly characters or words in input text, which can confuse the models but do not affect the labels judged by humans. In those works, such perturbations are not required to keep the semantics of original text. \cite{michel2019evaluationAdverSeq2Seq} argue that perturbations for \seqseq~tasks should minimize the change of semantic in input text but dramatically alter the meaning of outputs. \cite{cheng2020seq2sick} uses a sentiment classifier to verify whether sentiments of the original utterances is preserved after perturbation while we use crowdsourcing to ensure the meaning remains. 
Adversarial examples in semantic parsing cannot simply borrow from prior work because a parser does not make errors if it generates a different and correct LF when there is any subtle change in input text leading to change of semantics. In contrast, adversarial examples are supposed to cause parsing errors. Thus, adversarial examples w.r.t. meaning-changing perturbations are not well defined.

There are two types of methods in generating adversarial examples. The white-box methods~\cite{Nicolas2016Crafting,Javid2017White,athalye2018robustness} assume that the attacker have direct access to model details including their parameters, but the black-box methods assume that attackers have no access to model details except feeding input data and getting outputs from models~\cite{Gao2018Black-box,guo2019simple,Alvin2018Meta,Matthias2018comparing}.
\paragraph{Adversarial Training}
Adversarial training aims at improving the robustness of machine learning models against adversarial examples~\cite{goodfellow2014explaining,miyato2016adversarial,li-etal-2018-generating}. One line of research is to augment the training data with the adversarial examples. However, \newcite{javid2018character} points out that adversarial training may cause the model oversensitive to the adversarial examples. 
The other approach is to increase model capacity that may improve model's robustness \cite{Madry2018TowardsDL}. More techniques regarding adversarial defense can be found in the recent survey~\cite{Towards2019surveysemanticparsing}. In semantic parsing, data augmentation methods~\cite{jia2016datarecombination,guo2018question} are proposed only to improve performance of models on examples without perturbation but not to improve their robustness. 

\section{Evaluation Framework for Robust Semantic Parsing}
A robust semantic parser aims to transduce all utterances with or without meaning-preserving perturbations into correct LFs. Formally, let $\mathbf{x} = x_1, \dots, x_{|\mathbf{x}|}$ denote a natural language utterance, and $\mathbf{y} = y_{1}, \dots, y_{|\mathbf{y}|}$ be its LF, a semantic parser estimates the conditional probability (denoted by $p(\mathbf{y} | \mathbf{x})$) of an LF $\mathbf{y}$ given an input utterance $\mathbf{x}$. A robust parser's predictions are invariant to all $\mathbf{x}'$ that are generated from $\mathbf{x}$ by meaning-preserving perturbations.

For any $(\mathbf{x}, \mathbf{y})$, let $\sperturb$ denote the set of \textit{all} meaning-preserving perturbations of $\mathbf{x}$ that is parsed to $\mathbf{y}$:
\begin{equation}
    \sperturb = \{(\hat{\mathbf{x}}, \mathbf{y}) : \hat{\mathbf{x}} \in B(\mathbf{x}) \land o(\hat{\mathbf{x}}) = \mathbf{y} \}
\end{equation}
\noindent where $B(\mathbf{x})$ denotes the set of \textit{all} allowed perturbations of $\mathbf{x}$, and $o(\hat{\mathbf{x}})$ is an ideal parser that maps an utterance to its LF.  

A set of meaning-preserving perburbed examples $\Dperturb$ w.r.t.~a corpus $D$ is the union of all $\sperturb$ created from $D$. 

An adversarial example w.r.t.~a semantic parser is an utterance-LF pair, which is in $\sperturb$ and is parsed into an incorrect LF by that parser. A set of adversarial examples w.r.t.~a $\sperturb$ and a parser $f(\mathbf{x})$ is obtained by:
\begin{small}
\begin{equation}
   S_{\text{adv}} (\mathbf{x}, \mathbf{y}) =  \{(\hat{\mathbf{x}}, \mathbf{y}) : f(\hat{\mathbf{x}}) \neq \mathbf{y}, \forall (\hat{\mathbf{x}}, \mathbf{y}) \in \sperturb \} 
   \label{eq:adv}
\end{equation}
\end{small}
Subsequently, an adversary set w.r.t.~a semantic parsing corpus $D$ is created by taking the union of all adversary sets created from each example in $D$.

In the following, we present an evaluation framework for robust semantic parsing, which consists of an evaluation corpus, a set of evaluation metrics, and the corresponding toolkit for evaluating any new parsers.

\subsection{Construction of the Evaluation Corpus}
\label{sec:corpus}
We construct the evaluation corpus in a scalable manner by combining the existing semantic parsing benchmark corpora. Each of such corpora will be referred to as a \textit{domain} in the whole corpus. There are a train set, a validation set, and a standard test set in each domain. We perturb the examples in each test set to build a \textit{meaning-preserving test set} for each domain. More specifically, each example in a meaning-preserving test set is a perturbed utterance paired with its LF before perturbation. In the following, we detail each perturbation method and how we apply the crowdsourcing to remove meaning-changing ones. 

\subsubsection{Meaning-Preserving Perturbations}
\label{sec:perburb}
\vspace{-1.1em}
\begin{table}[ht!] 
\centering 
{\resizebox{\columnwidth}{!}{%
\begin{tabular}{|l | l|} 
\hline\hline 
\textbf{Perturbations} & \textbf{Meaning-preserving Examples} \\ 
\hline 
Insertion & \textcolor{red}{in} what state is the largest in population\\
Deletion & what state is \st{the} largest in population\\
Substitution (f) & what state is the largest \textcolor{red}{among} population \\
Substitution (nf) & what state is the \textcolor{red}{most}  in population\\
\hline\hline
Back Translation & state with the largest population \\
Reordering & the largest population is in what state \\
\hline 
\end{tabular}
}
}
  \vspace{-2mm}
  \caption{The meaning-preserving examples of the original utterance ``what state is the largest in population".}
  \label{tab:adv_examples}
    \vspace{-2mm}
\end{table}
Given an utterance, we perturb it by performing four different word-level operations and two sentence-level operations, \textit{respectively}. Table \ref{tab:adv_examples} lists the examples of generated meaning-preserving examples categorized according to the respective generation methods. More details are given below.


\paragraph{Insertion} 
Given an utterance $\mathbf{x}$, we randomly select a word position $t' \in \{1, \dots,|\mathbf{x}|\}$. A meaning-preserving example $\mathbf{x}'$ is created by inserting a function word at position $t'$. 

\paragraph{Deletion} We randomly remove a function word $x_t$ from $\mathbf{x}$.

\paragraph{Substitution (f)} Every function word in $\mathbf{x}$ is replaced with a random but different function word to generate a perturbed example.


\paragraph{Substitution (nf)} For every non-function word in $\mathbf{x}$, we apply the pretrained language model ELECTRA~\cite{clark2020electra} to select top-$k$ candidate words and exclude the original word. We generate a perturbed utterance for each valid candidate word. Since this method may generate utterances far from their original meaning, we subsequently filter those utterances by measuring their semantic similarity with the original ones. Specifically, we apply the SOTA sentence similarity model Sentence-Bert~\cite{reimers2019sentence} to compute similarity scores for each generated utterance resulting in only the $n$ highest scored utterances. 
\paragraph{Back Translation}
Inspired by~\cite{lichtarge2019corpora}, we revise utterances using back translation. We apply the Google translation API to translate utterances into a bridge language and then translate them back to English. Russian, French, and Japanese are used as the bridge languages to diversify and maximize the coverage of meaning-preserving perturbations. We select the best translation among all three translations for each original utterance according to Sentence-Bert's scores. 

\paragraph{Reordering} Similar to back translation, we reorder utterances by the SOTA reordering model SOW-REAP~\cite{goyal2020neural}. To increase the coverage, we follow the same strategy as \textit{Substitution (nf)} to generate an extended set of reordered utterances with multiple instances per input utterance. We use Sentence-Bert to encode sentences and select the top-$k$ best ones according to their cosine similarity scores with the original input. The reordered sentences share the same vocabulary as the original one except for the order of words.

\subsubsection{Filter Examples by Crowdsourcing}
As perturbation operations related to function words rarely alter meanings, we apply three crowdsourcing operations to the utterances perturbed, including \textit{Substitution (nf)}, \textit{Reordering}, and \textit{Back Translation}. For each perturbed utterance paired with the original utterance, three turkers at Amazon Mechanical Turk discern any semantic changes by choosing an option out of three --- \textit{the same}, \textit{different}, or \textit{not sure}. By default, any sentences uncomprehended by a human are regarded as \textit{not sure}. Finally, we keep only the ones that have the same meaning agreed by at least two turkers. After crowdsourcing, we keep 83\%, 82\% and 61\% of the generated utterances for \textit{Substitution (nf)}, \textit{Back Translation} and \textit{Reordering}, respectively.

\subsection{Evaluation Metrics}
Our framework assesses the performance of a semantic parser w.r.t.~standard accuracy and robustness metrics. Those robustness metrics indicate how well a semantic parser resists to meaning-preserving perturbations and adversarial attacks. As adversarial training is widely used for adversarial defense and mostly applicable to any neural semantic parsers, this framework supports comparing a wide range of adversarial training methods w.r.t.~standard accuracy and robustness metrics.

A training set, a validation set, a standard test set, and a meaning-preserving test set are established in each domain. The first three sets are obtained from the original benchmark corpus, and the meaning-preserving test set is created using the methods described in Sec.~\ref{sec:corpus}. We will examine the meaning-preserving test set (denoted by $\Dperturb$) and its two subsets. We refer to the first subset as the robustness evaluation set (denoted by $\Reval$), where the counterparts before perturbation are parsed correctly. We refer to the second subset as the black-box test set (denoted by $\Battack$), where the loss of a parser to the examples is higher than their counterparts before perturbation. For each target parser, we consider four metrics, including \textit{standard accuracy}, \textit{perturbation accuracy}, \textit{robust accuracy}, and \textit{success rate of black-box attack}.

\paragraph{Standard accuracy} The most widely used metric on semantic parsing~\cite{dong2018coarse2fine,yin2018tranx} to measure the percentage of the predicted LFs that exactly match their gold LFs in a standard test set. 

\noindent \paragraph{Perturbation accuracy} Perturbation accuracy is formally defined as $n / |\Dperturb|$, where $n$ denotes the number of correctly parsed examples to their gold LFs in a meaning-preserving test set $\Dperturb$. 

\noindent \paragraph{Robust accuracy} Robust accuracy is calculated as $n / |\Reval|$, where $n$ denotes the number of examples that are parsed correctly by a parser in a robustness evaluation set $\Reval$. Compared to perturbation accuracy, robust accuracy measures the number of examples that a parser can parse correctly before perturbation but fails to get them right after perturbation.

\noindent \paragraph{Success rate of black-box attack} A black-box attack example is regarded as the one that increases the loss of a model after perturbation~\cite{zhang2020adversarial}. Here, the success rate of black-box attack is calculated as $n / |\Battack|$, where $n$ denotes the number of examples that are parsed \textit{incorrectly} by a parser in the black-box test set $\Battack$.
White-box attacks require model specific implementation to generate adversarial examples, thus we leave the corresponding evaluation to the developers of semantic parsers.

The four metrics are computed to evaluate the efficacy of an adversarial training method. We inspect whether the metrics increase or decrease post-training and to what degree. An effective adversarial training method is expected to find a good trade-off between standard accuracy and robust accuracy.

Last but not least, all evaluation metrics are implemented with easy-to-use APIs in our toolkit. Our toolkit supports easy evaluation of a semantic parser and provides source code to facilitate integrating additional semantic parsing corpora. 

\section{Experiments}
In this section, we present the first empirical study on robust semantic parsing. 
\subsection{Experimental Setup}
\paragraph{Parsers}
We consider three SOTA neural semantic parsers --- \seqseq~with attention~\cite{luong2015effective}, \coarsefine~\cite{dong2018coarse2fine}, and \tranx~\cite{yin2018tranx}. \coarsefine~is the best performing semantic parser on the standard splits of \Geo~and \Atis. \tranx reports standard accuracy on par with \coarsefine and employs grammar rules to ensure validity of outputs.

\paragraph{Datasets}
Our evaluation corpus is constructed by extending three benchmark corpora --- \Geo~\cite{zelle1996geoquery}, \Atis~\cite{atis1994}, and \Jobs. \Geo contains 600 and 280 utterance-LF pairs to express the geography information in the training and test set, respectively. \Atis consists of 4434, 491, and 448 examples about flight booking in the training, validation, and test sets, respectively. And \Jobs~includes 500 and 140 pairs about job listing in the training and test set, respectively. 

\begin{table*}[ht]
\centering
   \resizebox{0.98\textwidth}{!}{%
  \begin{tabular}[t]{|l|c c c|c c c|c c c|}
    \toprule
    \multirow{2}{*}{Generation Methods} &
      \multicolumn{3}{c|}{\Jobs} &
      \multicolumn{3}{c|}{\Geo} &
      \multicolumn{3}{c|}{\Atis} \\
      & \seqseq $\quad$  & \coarsefine $\quad$ & \tranx $\quad$ & \seqseq $\quad$ & \coarsefine $\quad$ & \tranx $\quad$ & \seqseq $\quad$ & \coarsefine $\quad$ & \tranx \\
      \midrule
    Insertion &87.09/84.98 & 84.40/77.77  &81.89/82.97  & 85.59/80.50  & 87.57/85.21  & 85.96/81.73  & 67.36/68.33  & 84.16/82.73  & 77.45/75.63\\
    Substitution (f) &  86.58/78.89&84.22/71.59 & 82.23/86.10 & 85.27/78.38 & 86.97/82.30 & 84.23/79.04&70.93/59.19  &86.06/68.22  &76.70/67.69  \\
    Deletion &  86.48/82.23& 84.16/81.46 & 83.80/75.79 & 85.31/82.30 &85.69/72.56  &85.69/72.56  & 67.82/63.06 & 83.14/73.23 & 79.65/60.83 \\

    Substitution (nf) & 80.34/78.63 & 79.05/76.92 & 82.90/83.33 & 85.71/64.95 & 87.37/70.93 &86.37/66.11  & 75.28/56.36 & 85.57/68.94 & 78.55/60.57 \\
    \hline
    Back Translation $\,$ & 87.66/44.15 & 84.41/46.10  & 87.01/58.44 &85.92/48.74 & 86.93/56.28  &85.42/41.13 & 75.91/56.38 & 86.55/63.30 & 81.23/58.36 \\
    Reordering & 89.53/39.53 & 84.88/48.83 & 90.69/67.44&90.07/29.78  & 90.78/48.93 &93.61/41.13 & 72.28/40.66 & 87.04/52.10 &79.51/47.89 \\
    \hline
    Overall(micro) &86.15/76.08  &83.64/72.35  & 82.70/78.81 &85.48/72.55  &87.15/78.04  & 86.01/72.44 &70.72/60.55  &85.13/71.79  & 78.52/64.89 \\
    Overall(macro)&  86.27/68.06& 83.52/67.11 & 84.75/75.67  &86.31/64.10 & 87.55/69.36 & 86.88/63.61 &  71.59/57.33& 85.42/68.08 & 78.84/61.82 \\
        \bottomrule

  \end{tabular}%
   }
     \vspace{-2mm}
     \caption{Standard/Perturbation accuracy of SOTA parsers, trained on the original training sets.}
  \label{tab:rq1}
  \vspace{-2mm}
\end{table*}


\paragraph{Adversarial Training Methods}
We apply three adversarial training methods to the parsers. We evaluate whether the three adversarial training methods could improve semantic parsers' robustness against the meaning-preserving examples generated by the word-level and sentence-level operations. The corresponding three adversarial training methods are as follows:
\begin{itemize}
  \item[] \textbf{Fast Gradient Method~\cite{miyato2016adversarial}} Fast Gradient Method (FGM) adds small perturbations to the word embeddings and train the semantic parsers with the perturbed embeddings. The perturbations are scaled gradients w.r.t.~the input word embeddings. 
  \item[] \textbf{Projected Gradient Descent~\cite{madry2017towards}} Projected Gradient Descent (PGD) adds small perturbations to the word embeddings as well. Instead of calculating a single step of gradients, PGD accumulates the scaled gradients for multiple iterations to generate the perturbations.
  \item[] \textbf{Meaning-preserving Data Augmentation}  Meaning-preserving Data Augmentation (MDA) augments the original training data with the meaning-preserving examples generated by the word-level and sentence-level operations. We randomly select 20\% of the original instances for each dataset and generate their corresponding meaning-preserving instances. Since there are six different operations, each original training set is augmented with six different meaning-preserving sets independently. 
\end{itemize}

\paragraph{Training Details}
For both supervised training and adversarial training, we set batch size to 20 for all parsers with 100 epochs. We follow the best settings of three parsers reported in~\cite{dong2018coarse2fine,yin2018tranx} to set the remaining hyperparameters. The best performing implementation of \seqseq is included in~\newcite{yin2018tranx}. 

\begin{table*}[ht]
\centering
  \resizebox{0.9\textwidth}{!}{%
  \begin{tabular}{|l|c c c|c c c|ccc|}
    \toprule
    \multirow{2}{*}{Generation Methods} &
      \multicolumn{3}{c|}{\Jobs} &
      \multicolumn{3}{c|}{\Geo} &
      \multicolumn{3}{c|}{\Atis} \\
      & \seqseq  & \coarsefine & \tranx  & \seqseq & \coarsefine  & \tranx & \seqseq  & \coarsefine & \tranx \\
      \midrule
    Insertion & 96.06 & 91.51  & 94.96  & 82.32  &  96.45  & 95.37 & 93.42  & 95.04 & 93.91 \\
    Substitution (f) & 90.66 & 84.77 & 90.51 & 87.96 & 94.51 & 83.41 & 78.02  & 77.05 & 75.22 \\
    Deletion & 94.64  & 96.33 & 97.18 &88.03  & 93.67 & 92.04  & 88.04 & 86.06 &8 7.03  \\
    Substitution (nf) & 97.87 & 97.29 & 97.93 & 78.71 & 80.79 &76.64 & 76.50 &79.88  &  76.56\\
    \hline
    Back Translation & 48.88 & 54.07 & 64.92  & 61.07 & 64.16 &57.22 & 73.06  & 79.88 & 76.56 \\
    Reordering & 42.85 & 57.53 & 73.06 & 38.21 & 53.90 & 44.96 & 53.75 & 58.82 & 57.63 \\
    \hline
    Overall(micro) & 87.79 & 86.03 & 92.21 & 83.08 & 89.06 & 85.05& 81.29  & 82.27  & 83.06 \\
    Overall(macro) & 78.50 & 80.08 & 86.43 & 72.31 & 80.58  &74.94 &  77.71 & 78.12  & 76.82 \\

    \bottomrule
  \end{tabular}%
  }
    \vspace{-2mm}
    \caption{Robust accuracy for three SOTA parsers in each domain.}
  \label{tab:rq2}
    \vspace{-2mm}
\end{table*}

\subsection{Results and Analysis}
We discuss experimental results by addressing the following research questions:

\paragraph{RQ1: How do the SOTA parsers perform on meaning-preserving test sets?}


All three SOTA semantic parsers are trained on each training set before they are tested on the corresponding standard test sets and meaning-preserving test sets. Besides the overall results on whole test sets, Table \ref{tab:rq1} reports accuracy on each example subset perturbed by the respective perturbation operation. The results on those subsets are further compared with the standard accuracy on the corresponding examples before perturbation.

As shown in Table \ref{tab:rq1}, SOTA semantic parsers suffer from significant performance drop in almost all meaning-preserving test sets compared to the results on standard test sets. The performance ranking among the three SOTA semantic parsers varies across different datasets. \coarsefine achieves the best performance on \Geo and \Atis, while \seqseq beats \coarsefine and \tranx on \Jobs. Although \coarsefine achieves better accuracy than \tranx on the standard test set of \Jobs, it falls short of \tranx on the meaning-preserving test set of \Jobs. A parser achieving higher standard accuracy does not necessarily obtain better perturbation accuracy against meaning-preserving perturbations than its competitors.

Our evaluation framework supports also in-depth analysis on the impact of different perturbation method on semantic parsers. All parsers are more vulnerable to sentence-level perturbations than word-level ones. Although reordering changes only word order in utterances, it leads to the lowest perturbation accuracy of all parsers on \Geo and \Atis. Among word-level perturbation operations, substitution of non-function words is more challenging than deletion and insertion of function words on \Geo and \Atis. On \Jobs and \Atis, even deletion or substitution of function words can impose significant challenges for the parsers. When we further investigate the deleted or replaced function words, such as in Table \ref{tab:adv_examples}, semantic parsers are not expected to rely on such information. As all perburbations in our meaning-preserving test sets do not change meanings of original utterances, it imposes new research challenges on how to make semantic parsers resist meaning-preserving perturbations as well as how to avoid overfitting on semantically insignificant words.

\begin{figure*}[htp]
\small
  \centering
  \begin{subfigure}[b]{0.3\textwidth}
  \centering
    \includegraphics[width=\textwidth]{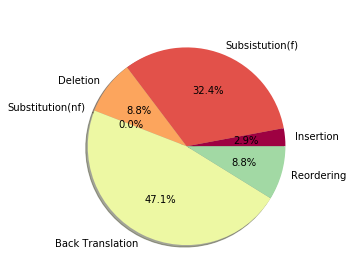}
    \caption{\Jobs}
  \end{subfigure}
  \hfill
  \begin{subfigure}[b]{0.3\textwidth}
    \centering
    \includegraphics[width=.9\textwidth]{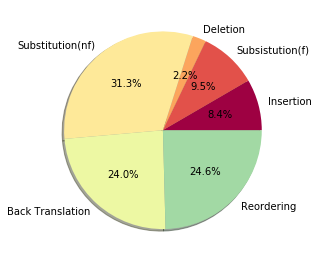}
    \caption{\Geo}
  \end{subfigure}
  \hfill
    \begin{subfigure}[b]{0.3\textwidth}
      \centering
    \includegraphics[width=.9\textwidth]{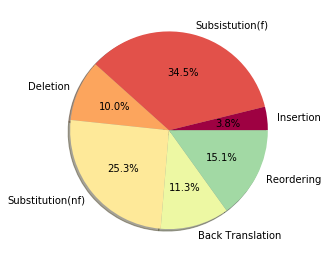}
    \caption{\Atis}
  \end{subfigure}
  
  \caption{Contributions of perturbation operations on the shared adversaries.}
  \label{fig:adv_set_intersection}
\end{figure*}

\paragraph{RQ2: What kind of perturbed examples particularly degrade parser performance?}




Although perturbation accuracy allows comparing parsers on the same test sets, it includes the examples that a parser fails to parse correctly both before and after the meaning-preserving perturbation. Robust accuracy focuses on the examples a parser parses successfully before perturbation but fails after that. We investigate all parsers trained without adversarial training in terms of this measure. As shown in Table \ref{tab:rq2}, \tranx is superior to the other two parsers on \Jobs, and \coarsefine is the clear winner on \Geo. This ranking of parsers is consistent with perturbation accuracy in the two domains. However, the differences among three parsers on \Atis are marginal, while \coarsefine achieves significantly superior perturbation accuracy in the same domain. \Atis is the domain with the most diverse paraphrases in natural language, \coarsefine cannot significantly outperform the other two parsers against meaning-preserving perturbations.

We further investigate adversary examples, which are defined in Eq. \ref{eq:adv}, for each parser in the meaning-preserving test sets. The shared adversarial examples among the parsers vary significantly across domains. More than 50\% of the adversarial examples are shared among the parsers on \Atis, but only less than 25\% of adversaries are shared on \Jobs. Fig.~\ref{fig:adv_set_intersection} illustrates which perturbation operation contributes to the intersection of the adversary set from different parsers. Back translation contributes almost half of the shared perturbed examples on \Jobs, while the proportion of word-level perturbation operations reaches nearly 70\% on \Atis. Although reordering and back translation impose the most difficult challenges for parsers, they cannot generate a significant number of valid adversarial examples. In contrast, it is relatively easy to resist a word substitution attack, but this operation can be easily applied to generate a large number of adversaries. 

\begin{table}[ht!] 

\centering 
\resizebox{\columnwidth}{!}{%
\begin{tabular}{|l | l|} 
\hline
\hline
\multicolumn{2}{|c|}{Insertion}  \\
\hline 
Utterance & show me the nonstop flight and fare from toronto to st. petersburg \\
Adv. Utterance & show me the nonstop flight and fare \textcolor{red}{from} from toronto to st. petersburg \\ 
\hline 
Ground Truth & ( lambda \$0 e ( and ( flight \$0 ) ( nonstop \$0 ) ( from \$0 toronto:ci ) ( to \$0 st\_petersburg:ci ) ) )\\
\hline
\seqseq & ( lambda \$0 e ( exists \$1 ( and ( flight \$0 ) ( nonstop \$0 ) ( from \$0 toronto:ci ) ( to \$0 st\_petersburg:ci ) ( = ( fare \$0 ) \$1 ) ) ) ) \\
\coarsefine & ( lambda \$0 ( exists \$1 ( and ( flight \$0 ) ( nonstop \$0 ) ( from \$0 toronto:ci ) ( to \$0 st\_petersburg:ci ) ( = (fare \$0 ) \$1 ) ) ) ) \\
\tranx & ( lambda \$0 e ( exists \$1 ( and ( flight \$0 ) ( nonstop \$0 ) ( from \$0 toronto:ci ) ( to \$0 st\_petersburg:ci ) ( = ( fare \$0 ) \$1 ) ) ) ) \\
\hline
\multicolumn{2}{|c|}{Substitution(f)} \\
\hline 
Utterance &what is the capital of the state with the largest population density \\
Adv. Utterance & what \textcolor{red}{are} the capital of the state with the largest population density \\ 
\hline 
Ground Truth & ( capital:c ( argmax \$1 ( state:t \$1 ) ( density:i \$1 ) ) )\\
\hline
\seqseq & ( lambda \$0 e ( and ( capital:t \$0 ) ( loc:t \$0 ( argmax \$1 ( state:t \$1 ) ( density:i \$1 ) ) ) ) ( density:i \$0 ) )\\
\coarsefine & ( argmax \$0 (state:t \$0 ) (density:i \$0 ) )  \\
\tranx & ( lambda \$0 e ( and ( capital:t \$0 ) ( loc:t \$0 ( argmax \$1 ( state:t \$1 ) ( density:i \$1 ) ) ) ) ) \\
\hline
\multicolumn{2}{|c|}{Delection(f)} \\
\hline 
Utterance &what is the capital of the state with the largest population density \\
Adv. Utterance &what is the capital \st{of} the state with the largest population density \\ 
\hline 
Ground Truth & 	( capital:c ( argmax \$1 ( state:t \$1 ) ( density:i \$1 ) ) )\\
\hline
\seqseq & ( argmin \$0 ( capital:t \$0 ) ( density:i \$0 ) )\\
\coarsefine & ( argmax \$0 (capital:t \$0 ) (population:i \$0 ) )  \\
\tranx & ( argmax \$0 ( capital:t \$0 ) ( density:i \$0 ) ) \\
\hline
\multicolumn{2}{|c|}{Substitution(nf)} \\
\hline
Utterance & what are the major river in ohio \\
Adv. Utterance & what are the \textcolor{red}{main} river in ohio \\ 
\hline 
Ground Truth & ( lambda \$0 e ( and ( major:t \$0 ) ( river:t \$0 ) ( loc:t \$0 ohio ) ) )\\
\hline
\seqseq & ( lambda \$0 e ( and ( river:t \$0 ) ( loc:t \$0 ohio ) ) ) \\
\coarsefine & ( lambda \$0 ( and (river:t \$0 ) (loc:t \$0 ohio ) ) )  \\
\tranx & ( lambda \$0 e ( and ( river:t \$0 ) ( loc:t \$0 ohio ) ) )\\
\hline
\multicolumn{2}{|c|}{Back Translation} \\
\hline
Utterance & list job using sql \\
Adv. Utterance & what kind of work do you have with sql\\ 
\hline 
Ground Truth &job ( ANS ) , language ( ANS , 'sql' ) \\
\hline
\seqseq & job ( ANS ) , application ( ANS , 'sql' ) \\
\coarsefine & job ( ANS ) , language ( ANS 'sql' ) , language ( ANS 'sql' )   \\
\tranx & job ( ANS ) , language ( ANS 'sql' ) , language ( ANS 'sql' )  \\
\hline
\multicolumn{2}{|c|}{Reordering} \\
\hline
Utterance & what is the lowest point in mississippi \\
Adv. Utterance & in s0 , the lowest point in mississippi\\ 
\hline 
Ground Truth & ( lambda \$0 e ( loc:t ( argmin \$1 ( and ( place:t \$1 ) ( loc:t \$1 mississippi:s  ) ) ( elevation:i \$1 ) ) \$0 ) ) \\
\hline
\seqseq & ( argmin \$0 ( and ( place:t \$0 ) ( loc:t \$0 mississippi:s  ) ) ( elevation:i \$0 ) ) \\
\coarsefine & ( lambda \$0 ( loc:t mississippi:s  \$0 ) ( loc:t (argmin \$1 ( and ( place:t \$1 ) ( loc:t \$1 mississippi:s  ) ) ( elevation:i \$1 ) ) \$0 ) )   \\
\tranx & ( argmin \$0 ( and ( place:t \$0 ) ( loc:t \$0 mississippi:s  ) ) ( elevation:i \$0 ) ) \\
\hline\hline 
\end{tabular}
}

  \caption{Adversarial examples shared by all parsers.}
  \label{tab:display_examples}
\end{table}
We hand picked representative adversarial examples shared by all parsers and list them in Table \ref{tab:display_examples}. The errors made by substitution (f) and insertion often show a sign of overfitting due to the fact that the parsers learn dependencies between non-essential features and predicates. Substitution (nf) sometimes causes predicates to be missing after replacing a word with its synonym. The errors caused by deletion, back translation, and reordering are more diverse, including adding wrong predicates or missing predicates.  


\begin{figure*}[htp]
\small
  \centering
  \begin{subfigure}[b]{0.3\textwidth}
  \centering
    \includegraphics[width=\textwidth]{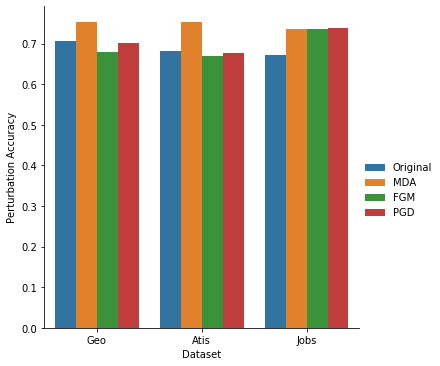}
    \caption{Perturbation accuracy.}
  \end{subfigure}
  \hfill
  \begin{subfigure}[b]{0.3\textwidth}
    \centering
    \includegraphics[width=\textwidth]{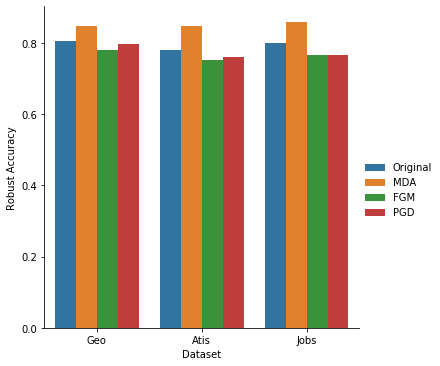}
    \caption{Robust accuracy.}
  \end{subfigure}
  \hfill
    \begin{subfigure}[b]{0.3\textwidth}
      \centering
    \includegraphics[width=\textwidth]{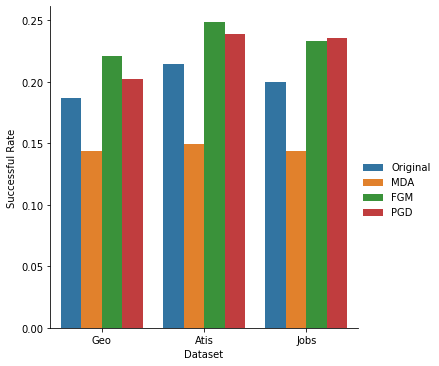}
    \caption{Success rate of black-box attack.}
  \end{subfigure}
  
  \caption{Robustness metrics of \coarsefine before and after adversarial training.}
  \label{fig:metrics_adv_training}
\end{figure*}

\begin{table*}[ht]
\centering
   \resizebox{0.9\textwidth}{!}{%
  \begin{tabular}[t]{|l|l l l|l l l|l l l|}
    \toprule
    \multirow{2}{*}{Generation Methods} &
      \multicolumn{3}{c|}{\Jobs} &
      \multicolumn{3}{c|}{\Geo} &
      \multicolumn{3}{c|}{\Atis} \\
      & \seqseq $\quad$ & \coarsefine $\quad$ & \tranx $\quad$ & \seqseq $\quad$ & \coarsefine $\quad$ & \tranx $\quad$ & \seqseq $\quad$ & \coarsefine $\quad$ & \tranx \\
      \midrule
    
    No adv. train &87.14  & 83.57  & 87.14  &83.92  &86.78   & 86.07  & 73.88  & 86.16 & 79.91\\
    \hline
    FGM $\,$&87.31 & 86.12 & 91.34  & 83.89 & 86.35 & 84.01  &81.00 & 79.81 & 81.72   \\
    PGD &82.84 & 86.18 &92.93  & 84.56 &86.35 &84.01 & 79.96 &79.94  &81.52 \\
    MDA(all)  &87.74 &86.16  &88.57  &  82.89& 86.73&87.11 &74.75  &80.18  &80.60 \\

    \hline
    MDA(Insertion) & 89.14 &86.16  &88.85  & 81.78 & 87.14 &87.64 & 67.16 & 85.27& 78.75 \\

    MDA(Substitution (f)) & 87.71 & 85.25 &88.85  & 83.21 &86.79  &87.35 & 76.51 & 78.08& 85.67 \\

    MDA(Deletion) & 87.28 &90.14 & 85.33 & 83.21 & 86.79& 87.35 &76.87  &85.62   & 78.52 \\
    MDA(Substitution (nf)) & 87 & 80.68 &90.14  & 83.85 &86.79 &86.85 & 76.11 & 62.58  & 85.58 \\
    \hline
    MDA(Back Translation) $\,$&88.42 &87.53  & 88.57  & 82.14 & 86.71 & 86.64  &76.20 & 84.73 & 77.36   \\
    MDA(Reordering) &86.91 & 87.21 & 89.71 &83.21  &86.21 &86.85 &75.66  & 84.82 & 77.72\\
    \bottomrule
  \end{tabular}%
   }
          \vspace{-2mm}
  \caption{Standard accuracy of adversarial training, which is trained with adversarial training methods and evaluated on the standard test sets. MDA(all) stands for MDA with all six perturbation operations.}
  \label{tab:rq4}
       \vspace{-2mm}
\end{table*}

\paragraph{RQ3: How effective are those widely used adversarial training methods for semantic parsers?}

We applied FGM, PGD, and MDA to train the three parsers on the respective training sets. In the case of MDA, we consider all automatic perturbation operations to generate training examples, and each operation adds the same amount of training examples to the original train sets. Fig.~\ref{fig:metrics_adv_training} illustrates the results of \coarsefine w.r.t.~different evaluation metrics. MDA always improves the three evaluation metrics, despite that the generated training examples are not checked by humans, resulting in a significant number of meaning-changing examples. FGM and PGD hardly work except on \Jobs in terms of perturbation accuracy. Since both methods are white-box methods, the results indicate the gradient-based adversaries cannot effectively capture the meaning-preserving perturbations.

The biggest improvement made by MDA is the success rate of a black-box attack. The black-box attack examples are selected from the meaning-preserving examples that increase a parser's loss after perturbation. When we inspect the examples in the meaning-preserving test sets generated by different perturbation operation, the biggest improvement is from the ones perturbed by sentence-level operations and word substitution. All three parsers benefit from MDA significantly. The more vulnerable to black-box attack a parser is, the more improvement it can achieve after applying MDA. 

\paragraph{RQ4: How does each adversarial training method influence standard accuracy?}

\newcite{tsipras2018robustnessOdds} point out that there is a trade-off between standard accuracy and robust accuracy in most occasions. This finding is attributed to the presence of \textit{non-robust} features, which are highly predictive but incomprehensible for humans~\cite{ilyas2019adversarialNotBugs}. To verify the theory, after applying each adversarial training method to each parser, we compare the standard accuracy before and after adversarial training.

As shown in Table \ref{tab:rq4}, none of the three adversarial training methods consistently improve standard accuracy across different domains and parsers. \tranx does not have a significant performance drop regardless which adversarial training method is applied. It may be due to the fact that \tranx uses grammar to filter out invalid outputs.  

MDA cannot consistently improve parsers but also does not hurt parsers' performance in terms of standard accuracy. 
We conducted t-tests on the standard test sets to assess if MDA with different perturbation operations significantly improves accuracy. The results are negative so that the training examples generated by MDA at least do not hurt parsers' performance while increasing their robustness.

\paragraph{RQ5: How does our meaning-preserving data augmentation method compare with the data augmentation method proposed by \cite{jia2016datarecombination}?}
\cite{jia2016datarecombination} is one of the SOTA data augmentation methods for semantic parsing. In their work, they show that the augmented examples improve accuracy of predicting denotations. Although there are three methods proposed in \cite{jia2016datarecombination}, only the method of \textit{concatenating} two random examples as a new example can be applied in our case. We evaluated this augmentation method with all three parsers. No significant improvement of standard accuracy and robust accuracy is found in all three domains. We conjecture that this is due to the fact that in \cite{jia2016datarecombination} they report only improvement of accuracy of denotation matching, not the matching of LFs. In contrast, MDA can effectively reduce the harm of meaning-preserving perturbations.

%


\subsection{Conclusion}
We conduct the empirical study on robustness of neural semantic parsers. In order to evaluate robustness accuracy, we define first what are adversarial examples for semantic parsing, followed by constructing test sets to measure robustness using a scalable method. The outcome of this work is supposed to facilitate semantic parsing research by providing a benchmark for evaluating both standard accuracy and robustness metrics. 

\bibliographystyle{acl_natbib}
\bibliography{eacl2021}

\end{document}